\documentclass{artificial-life}
\usepackage[style=apa,natbib=true]{biblatex}
\bibliography{NAME}
\usepackage{amsmath}

\title{Behaviour diversity in a walking and climbing centipede-like virtual creature}

\auth{
    Emma Stensby Norstein \affil{1},
    Kotaro Yasui \affil{2},
    Takeshi Kano \affil{3},
    Akio Ishiguro \affil{2,5},
    Kyrre Glette \affil{1, 4,5}
}

\affiliations{
    \item Department of Informatics, University of Oslo, Norway
    \item Tohoku University, Sendai, Japan
    \item Future University Hakodate, Hakodate, Japan
    \item RITMO, University of Oslo, Norway 
    \item These authors share senior authorship.
}

\abs{
Robot controllers are often optimised for a single robot in a single environment. This approach proves brittle, as such a controller will often fail to produce sensible behavior for a new morphology or environment.
In comparison, animal gaits are robust and versatile.
By observing animals, and attempting to extract general principles of locomotion from their movement, we aim to design a single decentralised controller applicable to diverse morphologies and environments. 
The controller implements the three components 1) undulation, 2) peristalsis, and 3) leg motion, which we believe are the essential elements in most animal gaits. The controller is tested on a variety of simulated centipede-like robots. The centipede is chosen as inspiration because it moves using both body contractions and legged locomotion.
For a controller to work in qualitatively different settings, it must also be able to exhibit qualitatively different behaviors. We find that six different modes of locomotion emerge from our controller in response to environmental and morphological changes. We also find that different parts of the centipede model can exhibit different modes of locomotion, simultaneously, based on local morphological features.
This controller can potentially aid in the design or evolution of robots, by quickly testing the potential of a morphology, or be used to get insights about underlying locomotion principles in the centipede.
}

\keywords{Morphology, Control, Insect locomotion, Robotics}

\begin{document}

\coverpage

\section{Introduction}

Traditional approaches to robot control usually result in a controller that is tailored to a particular robot, and often to a particular environment or task. This type of controller, which is optimized for a specific morphology and environment, will often fail if the morphology is changed or damaged \citep{diff, adaptdamage1, adaptdamage2}, or if the robot encounters a novel environment. \citep{adaptterrrain1, adaptterrrain2}
In contrast, biological organisms show an impressive versatility and adaptability.\citep{robustinsect1, robustinsect2}
Animals constantly and efficiently adapt to growing bodies, new places, and injuries. They exhibit diverse behavior, and quickly adapt to situations they encounter for the first time.
This suggests that there is an underlying control mechanism in animals that is capable of adapting to changes in both the body and environment. 
Taking inspiration from this, we seek to decouple robot control from robot morphologies and environments. We want to design a general  robot controller that exhibits diverse modes of locomotion in response to a diverse array of morphologies and environments, with little parameter tuning and no modification.

\begin{figure}[h!]
    \centering
    \includegraphics[width=\textwidth]{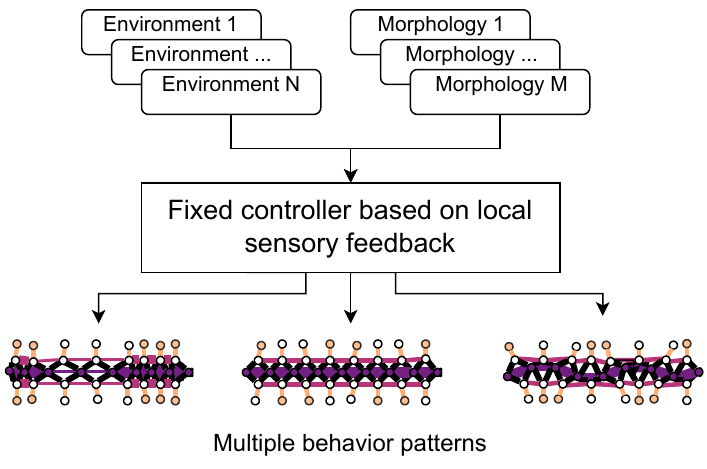}
    \caption{Illustration of the experiment setup. The controller adapts to the environment and morphology of the centipede model based on proprioceptive feedback, creating a variety of gait patterns. The controller requires no training or retuning upon encountering a new environment or morphology.}
    \label{fig:concept}
\end{figure}

A general robot controller could advance several fields within robotics. Used on a traditional hand designed robot it would likely be robust to both damage and transfer from simulation to reality \citep{simulationER, realitygap}.
A controller that can adapt to its morphology and environment could adapt to a damaged robot as if it was a new morphology, and to reality as if the physical robot and physical environment were a new morphology and a new environment.
A general controller could also be useful within the field of morphology evolution. In this field the shape of a robot is optimized using an evolutionary algorithm, often alongside the robot controller \citep{sims, ER}. When evolving morphologies it is common to get stuck in local optima due to a lack of time for optimizing controllers to each new morphology that appears \citep{diff}. 
While promising approaches such as protection of new morphologies to allow time for controller optimization~\citep{scalable}, and time-efficient lifetime learning approaches~\citep{van_diggelen_comparing_2024} have been explored, the computational overhead is still significant.
With a general controller the potential of a morphology could be evaluated quickly, leading to less time spent exploring useless robot designs. 
For these reasons, we believe morphology and environment adaptive control to be a fruitful research direction.

The idea of an adaptive controller, that can control multiple morphologies, has been investigated both in the field of bio-inspired robotics \citep{biorobotics} and in evolutionary or modular robotics \citep{ER, modularrobots}. Within modular robotics the focus is often on reinforcement learning methods. \citep{mertan_modular_2023, reinforcementmanymorph1, reinforcementmanymorph2, reinforcementmanymorph3, reinforcementmanymorph4} These approaches usually require a lot of computational resources for training neural networks, and can be difficult to analyse due to large black box models. However, their benefit is that they can work without the designer needing a deep understanding of the complex kinematic system underlying the gaits. The bio-inspired and reinforcement learning approaches have also been combined by \citet{manyquadrupeds}, who combines central pattern generators and reinforcement learning to create a policy that can control a diverse set of quadruped morphologies.

Our approach to designing a general controller is that of bio-inspired robotics. We leverage mechanics found in animals, and try to extract general principles of locomotion from their movement patterns. If we can understand why animal behavior has adapted in a particular way, in response to a morphological or environmental setting, we can use this information to design more efficient and general robot controllers. A challenge in this endeavor is to find the correct balance between capturing diverse behaviors, and keeping the control principles simple. Animals exhibit a diverse range of behaviors, but a lot of animal gaits can be explained in terms of three essential elements, 1) undulation, 2) peristalsis, and 3) leg motion. We decide to focus on these three types of movement in our controller. 

In previous studies the locomotion principles of many different animals have been explored. Some notable examples are studies of the gaits of centipedes \citep{myriapod, centipedecontroller}, caterpillars \citep{caterpillar}, earthworms \citep{earthworm}, and brittle stars. In their study on brittle stars, \citet{brittlestar} demonstrate that their brittle star controller can automatically adapt its gait to create functional locomotion for brittle star robots where up to four of its five arms are amputated. A lot of the work in the area of adaptive control focuses on adaptation to damage, as this is a case where a general controller would be useful. Hayakawa et al. automatically generate gaits for various configurations of one legged robotic modules, and show that their gait generation is robust to damage in the morphology \citep{legformations}.  Miguel-Blanco et al. use a neural CPG-based model to generate gaits for myriapod robots with varying numbers of legs, and show that the controller automatically adapts to leg malfunctions \citep{myriapod}. Most studies in the field is focused on a single mode of locomotion. However, \citet{manymorphologies} have created at general framework to design gaits that combine the slithering locomotion of snakes with legged locomotion. \citet{manymorphologies}. extend the Hildebrand gait formulation \citep{hildebrand} to generate gaits for a quadruped, hexapod, myriapod and a sidewinder robot. 

In this paper we choose to focus on producing locomotion for centipede-like morphologies. The centipede has a long snake- or earthworm-like body, but it also has legs. This means that we can explore undulatory and peristaltic body motion, while at the same time exploring legged locomotion and body-limb coordination. The centipede displays a wide variety of behaviors, including undulatory gaits with direct and retrograde waves \citep{gaitswitch}, and peristalsis \citep{japanesecentipede}. Different species of centipede exhibit different patterns. Some species use only direct wave gaits, some use only retrograde wave gaits, and some use both. We believe this difference in appearance of gaits may be the result of an underlying control principle adapting to the different centipede anatomies. This adaptation would be a property we want to reproduce in our morphology adaptive controller. In addition to this there are some species of centipede where the number of segments in the centipede’s body changes over the course of its life \citep{Lewis1981}, indicating that the centipede controls its body in a decentralized manner that is independent of the body length. These properties make centipedes an excellent starting point for our study.

We have implemented and tested a controller for centipede locomotion, as shown in figure \ref{fig:concept}. The controller is inspired by the centipede controller made by \citet{centipedecontroller}. However, the controller is extended to modulate the leg motion based on peristaltic contractions in the trunk. This adds additional functionality in the form of more diverse modes of locomotion. While the controller made by \citet{centipedecontroller} was only capable of producing undulatory gaits, our new controller is capable of producing both undulatory and peristaltic gaits, and the model automatically switches between these gaits based on environment and morphology. Because the controller is fully decentralized, a centipede model can exhibit different patterns in different parts of its body.
The controller is tested on simulated virtual creatures with morphologies resembling centipedes, and the centipede models can both climb and walk.
We investigate the controller's adaptability to changes in both morphology and environment. The adaptation to morphology is explored by changing the length of the centipede model's legs, and the mass of its trunk, in a walking scenario. The adaptation to environment is explored by changing the incline of a pole in a climbing scenario.

We show that the controller can exhibit six different gaits depending on body contraction amplitude, morphology and environment.
\footnote{A video demonstrating the patterns can be found at \url{https://www.mn.uio.no/ifi/personer/vit/emmaste/centipede_patterns.mp4}} The six gaits are 1-2) retrograde wave patterns with high/low body undulation, 3-4) direct wave patterns with high/low body undulation, and 5-6) contralateraly in-phase patterns with high/low peristalsis. We also show that the locomotion mode automatically switches from a retrograde to a direct pattern based on leg length or trunk mass, and from an undulatory to a contralateraly in-phase pattern based on environment.  This demonstrates that our controller adapts automatically to both morphology and environment. The controller requires no training, and no tuning to new morphologies and environments. Instead it converges to a gait during the first part of the evaluation in the simulation environment. To demonstrate the flexibility of our controller, we also show a mixed leg length morphology where two gait patterns co-exist within a single body.

\section{Method}

We have created a controller for locomotion of centipede-like virtual creatures. 
The controller exhibits locomotion based on three essential elements found in animal gaits, 1) undulation, 2) peristalsis, and 3) leg motion. 
The controller is evaluated through simulation of a centipede model, which we implemented in Unity 3D, using Unity ML-Agents \citep{unityMLagents}. The experiments are conducted in two different scenarios, one where the centipede is walking, and one where it is climbing. In the walking scenario the leg length and body mass
is varied to evaluate the controller’s adaptability to different morphologies. In the climbing environment the angle at which the centipede climbs is varied to evaluate the controller's adaptability to environmental changes. The following sections describe 1) the centipede model, 2) the two simulation scenarios, 3) details about the Unity simulation environment, 4) the controller architecture, and 5) categorisation of the emerging gait patterns.

\subsection{Centipede model}
\begin{figure}[tp]
    \centering
    \includegraphics[width=\textwidth]{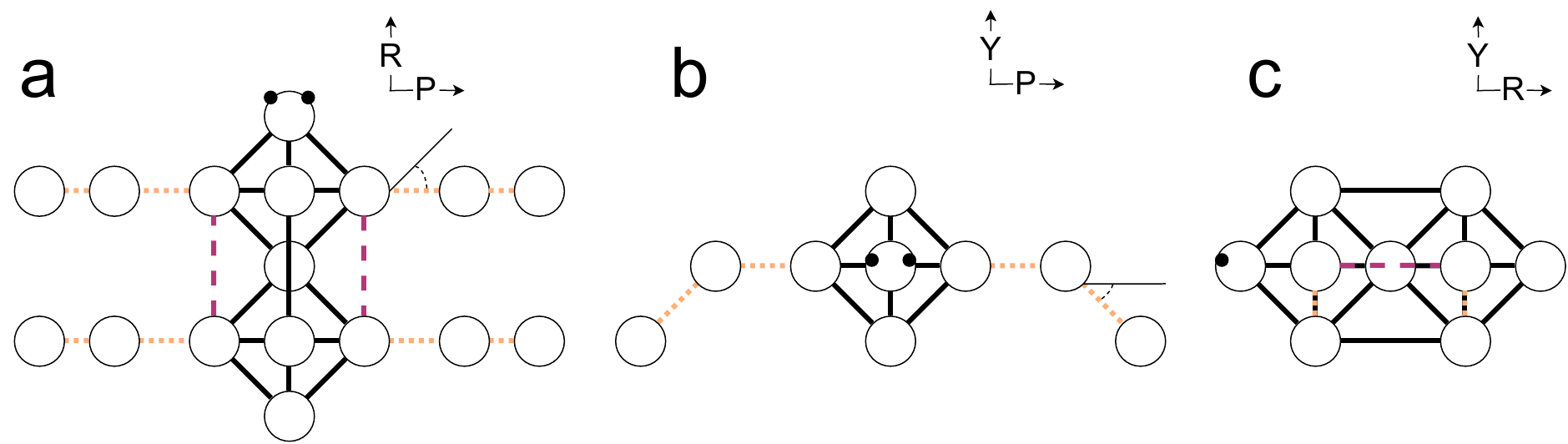} 
    \caption{
    The centipede model in a) top view, b) front view, and c) side view.
    The solid black lines are passive spring-damper connections, the pink dashed lines are linear actuators, and the orange dotted lines are rotational actuators. The axes denote the (Y) yaw, (P) pitch, and (R) roll axes of the model.}
    \label{fig:centipede_model}
\end{figure}

The centipede model consists of a string of segments, with one pair of legs per segment. Each segment of the model has ten mass points, which are connected with spring-damper systems. The connections can be either passive or active, and the active connections can be either linear or rotational actuators.
The power that can be exerted by the connections is determined by their spring and damper values, which correspond to the $k_p$ and $k_d$ values of a PD-controller. The internal physical model of a segment is visualized in Figure \ref{fig:centipede_model}. The sizes and weight of the centipede model are given without units. This is because they are units of length and weight in Unity, which do not correspond to real world values (see section \ref{section:unity_details} for more details).

Six of the mass points are arranged in an octahedron shape, and make up the trunk segment. The width of the octahedron is 1. The front and back mass points in the trunk segment are respectively shared with the segment in front of and behind the segment. The remaining four mass points make up the legs. 

The spring-damper joints connecting the mass points are sorted into five categories with different spring damper values. The five categories are 1) the active rotational joints connecting the feet to the knee, 2) the active rotational joints connecting the knees to the trunk, 3) the passive connections making up the octahedron shape within a trunk segment, 4) the passive connections connecting two trunk segments to each other on the back and belly of the model, and 5) the active linear joints connecting two trunk segments to each other on the left and right of the model.

\paragraph{Body parameters} The spring and damper values we used for the body joints are summarized in the table below.
The values were found through hand tuning, which took place during the design process of the simulation environment and model.

\begin{center}
\label{tab:pd_parameters}
\begin{tabular}{|l|c|c|c|c|}
    \hline
    Connection & Linear spring & Linear damper & Rotational spring & Rotational damper \\
    \hline
    1, Foot to knee & 4500 & 5 & 7000 & 40 \\
    2, Knee to trunk & 1750 & 20 & 8500 & 0 \\
    3, Body internal & 5000 & 300 & 600 & 0 \\
    4, Body passive & 5500 & 10 & 5000 & 20 \\
    5, Body active & 1500 & 20 & 6000 & 30 \\
    \hline
\end{tabular}
\end{center}

\paragraph{Morphology variation}

In experiment 1 the length of the legs is varied. The leg segment connecting the knee to the trunk is always 30\% of the total leg length, while the leg segment connecting the foot to the knee is always 70\% of the total length. In experiment 2 the weight of the trunk is varied. In experiment 4 the front half of the legs are long and the back half of the legs are short. All other morphology parameters are constant, and the morphology parameters for each experiment are summarized in the table below. 

\begin{center}
\label{tab:centipede_parameters}
\begin{tabular}{|l|c|c|c|c|}
    \hline
    Centipede model part & Experiment 1 & Experiment 2 & Experiment 3 & Experiment 4 \\
    \hline
    Leg length & \textbf{1-8.75} & 3 & 1.5  & \textbf{1/3}\\
    Trunk width & 1 & 1 & 1 & 1\\
    Number of segments & 12 & 12 & 12 & 22\\
    Leg weight & 0.1 & 0.1 & 0.1 & 0.1\\
    Trunk segment weight & 0.3 & \textbf{0.06-3.66} & 0.3 & 0.3\\
    \hline
\end{tabular}
\end{center}

\subsection{Scenarios}
\begin{figure}[tp]
    \centering
    \includegraphics[width=\textwidth]{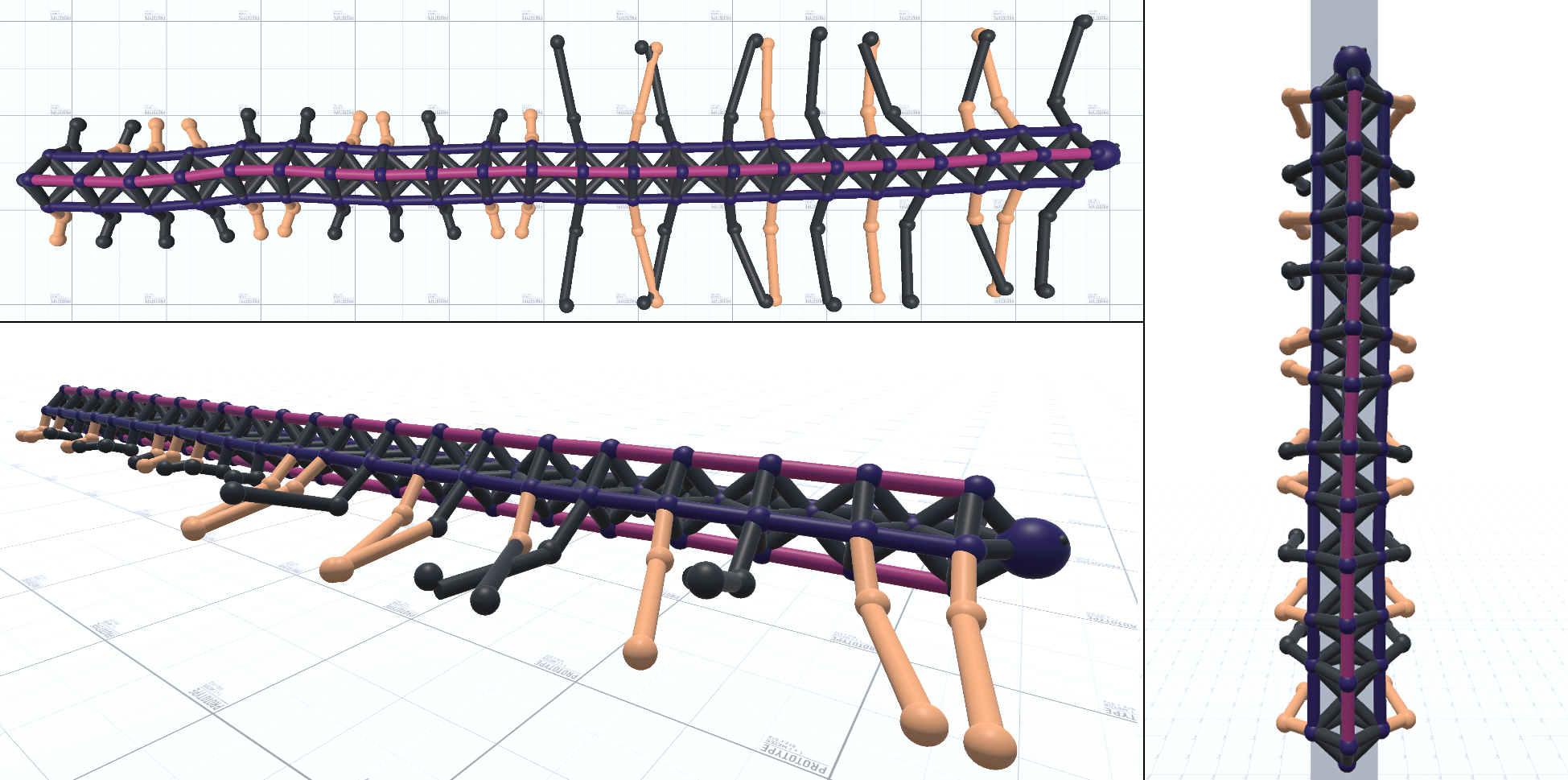}
    \caption{The centipede model is first tested while walking on a flat plane (left), then while climbing up a pole (right).}
    \label{fig:scenarios}
\end{figure}

We test our controller in two scenarios. In the first scenario the centipede is walking on a flat plane. In the second scenario it is climbing up a pole. Images of the two scenarios can be seen in Figure \ref{fig:scenarios}. 
In the climbing scenario the neutral angle of the legs is increased by 115 degrees, so that the legs fold around the pole, allowing the centipede to hold on to the pole. The pole has a square shape, and the width of the pole is 1.12, and the incline of the pole is varied between 0 and 93 degrees. In both scenarios the initial phase of each leg is a random value between 0 and 2$\pi$.

\subsection{Simulator}
\label{section:unity_details}
The simulation environment is created in Unity \citep{unityMLagents}, using the Nvidia PhysX 4.1 rigid body physics simulator. We used an accuracy of up to 255 iterations per physics steps in the physics solver. The settings for the simulation can be found in Table \ref{tab:physics_parameters}.

\begin{center}
\label{tab:physics_parameters}
\begin{tabular}{|l|r|}
    \hline
    Iterations & $255$ \\
    Gravity & $-24.525$ \\
    Coulomb friction coefficient & $5$ \\
    \hline
\end{tabular}
\end{center}

\subsection{Controller}
The controller consists of three parts 1) the body controller, which controls the contractions of the left and right linear joints connecting two trunk segments, 2) the leg controller, which controls the phase and movement of the legs, and 3) the foot controller, which moves the feet up and down based on the leg phase.

\begin{figure}[tp]
    \centering
    \includegraphics[width=\textwidth]{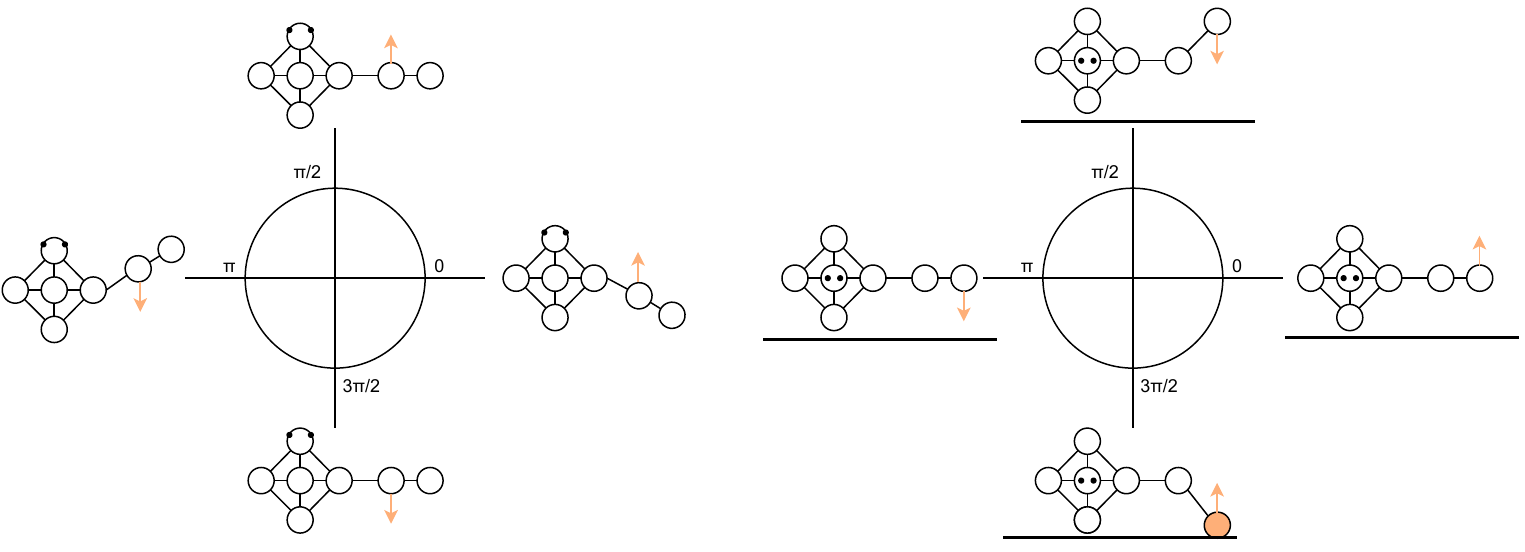} 
    \caption{The relationship between the leg phase and the position of the leg (left), and between the leg phase and the position of the foot (right). The arrow shows the direction that the leg or foot is heading. Note that the left diagram shows the model in top view, while the right diagram shows it in front view.}
    \label{fig:phase}
\end{figure}

\paragraph{Body controller} 
The body controller we use is designed by \citet{centipedecontroller}. The body controller sets the desired linear contraction of the two active joints connecting a trunk segment to the next segment. If only one of the two linear joints contract it will cause the trunk segment to bend, while if both sides contract it will cause the trunk segment to contract. The controller uses sensory information from the feet to determine the contractions. The desired length of a linear joint, $\bar{l}$, is found with 
the following equation:

\begin{equation}\label{eqn:bc}
\bar{l}^P_{i+\frac{1}{2}} = l_n - \beta\tanh(\gamma N_{i+1}^P),
\end{equation}

where $P$ can be either $L$ or $R$, which are the left and right side of the model, $i$ is the segment number indexed from tail to head ($i+\frac{1}{2}$ means that the joint connects segment $i$ and $i+1$), $l_n$ is the natural length of the joint, $\beta$ is a parameter that determines the amplitude of the contractions, $N$ is input from a pressure sensor at the foot, and $\gamma$ is a parameter that determines the sensitivity to $N$. The $\beta$ parameter is not tuned along with the other parameters. Instead it simulates a signal from the centipede brain that controls locomotion speed, and is varied in the experiments.

The sensory feedback N is calculated from the impact between the foot and the ground. To remove noise on the controller’s input signal, the average impact over the previous 20 time steps is used as the sensory feedback. This causes the joint to gradually contract over 20 time steps once the leg is in contact with the ground, and gradually release over 20 time steps once the contact stops. The window of 20 time steps was found by gradually increasing the window size until the signal became stable.

\paragraph{Leg controller}
The leg controller moves the legs around the yaw axis of the centipede.
The angles of the legs are controlled by oscillators. Figure \ref{fig:phase} shows the relationship between the oscillator phase and the position of the legs and feet. 
The phase of the oscillator is modulated based on proprioceptive sensory information from the body. We compare three variants of the controller, each with different proprioceptive feedback. The three variants receive feedback from 1) an angle detector, 2) a contraction detector, and 3) both the angle and contraction detectors. We call the three variants A, C and A+C. The equations for the angle detector $A_D$ and contraction detector $C_D$ are:

\begin{align}\label{eqn:AT}
& A_i^D = l^R_{i-\frac{1}{2}} - l^L_{i+\frac{1}{2}} + l^R_{i+\frac{1}{2}} - l^L_{i-\frac{1}{2}} \\
& C_i^D =  l^L_{i+\frac{1}{2}} + l^R_{i+\frac{1}{2}} - l^L_{i-\frac{1}{2}} -l^R_{i-\frac{1}{2}}
\end{align}

where $L$ and $R$ are the left and right side of the model, $l$ is the measured current length of a linear joint in the trunk 
, and $i$ is the segment number indexed from tail to head ($i+\frac{1}{2}$ means that the joint connects segment $i$ and $i+1$). 

The angle detector uses the difference between the contractions on the left and right side of two trunk segments to determine whether a leg is on the concave or convex side of a bend in the trunk. The contraction detector uses the difference between contractions above and below a trunk segment to determine whether a trunk segment is pulled forward or backward compared to the rest of the trunk.

The following equations show the phase evolution of the left leg and right leg for each of the three variants of the controller. Variant 1 of the controller is designed by \citet{centipedecontroller}. However, $A^D_{i}$ is computed differently from how it is computed by \citet{centipedecontroller}. 

Variant 1, with the angle detector, A:

\begin{align}\label{eqn:lcA}
\frac{d \theta^R_i}{d t} = \omega + \sigma_A A^D_{i} \cos(\theta^R_i)\\
\label{eqn:lcA_2}
\frac{d \theta^L_i}{d t} = \omega - \sigma_A A^D_{i} \cos(\theta^L_i)
\end{align}

Variant 2, with the contraction detector, C:

\begin{align}\label{eqn:lcT}
\frac{d \theta^R_i}{d t} = \omega - \sigma_T  C^D_{i} \cos(\theta^R_i + c\pi)\\
\label{eqn:lcT_2}
\frac{d \theta^L_i}{d t} = \omega - \sigma_T  C^D_{i} \cos(\theta^L_i + c\pi)
\end{align}

Variant 3, with both the angle and contraction detectors:

\begin{align}\label{eqn:lcAT}
\frac{d \theta^R_i}{d t} = \omega + \sigma_A A^D_{i} \cos(\theta^R_i) - \sigma_T  C^D_{i} \cos(\theta^R_i + c\pi)\\
\frac{d \theta^L_i}{d t} = \omega - \sigma_A A^D_{i} \cos(\theta^L_i) - \sigma_T  C^D_{i} \cos(\theta^L_i + c\pi)
\end{align}

where $t$ is the time, $i$ is the segment number indexed from tail to head, $\theta^R$ is the phase of the right leg, $\theta^L$ is the phase of the left leg, $\omega$ is the angular velocity, $\sigma_A$ is a parameter that determines the strength of the angle detector modulation, $\sigma_T$ is a parameter that determines the strength of the contraction detector modulation, and $c$ is a parameter that affects the phase difference between legs in an in-phase pattern (see Section \ref{sec:patterndetection} on detecting patterns).

The angle of the leg joint, $LEG$, is determined with the following equation:

\begin{equation}\label{eqn:lc}
LEG^P_i = \alpha_{LEG} \cos(\theta^P_i)
\end{equation}

where $P$ can be either $L$ or $R$, which are the left and right side of the model, $i$ is the segment number indexed from tail to head, $\theta$ is the phase of the leg, and $\alpha_{LEG}$ is the amplitude of the leg movement.

Intuitively we can say that modulation by the angle detector produces an undulatory gait pattern by pulling a leg to stance phase when the corresponding trunk segment is rotated towards the leg. The phase modulation in equation \ref{eqn:lcA} has the following behaviour when the trunk segment is rotated towards the right leg: The angular velocity of the right leg speeds up as it is approaching stance phase ($\frac{3\pi}{2}$), and slows down as it is moves away from stance phase. This pulls the phase of the right leg towards stance phase. When the trunk segment is rotated away from the right leg, the modulation is opposite, as the sign of $A^D$ will be the opposite. The leg will then be pulled towards swing phase instead. For the left leg, equation \ref{eqn:lcA_2}, the sign of the modulation term is negative, which means that the left leg will instead be pulled to stance phase when the trunk segment is rotated to the left, and to swing phase when the trunk segment is rotated to the right. This modulation, in summary, pulls the phase of each leg to a behavior where each leg steps when the trunk segment is rotated towards it, and swings when the trunk segment is rotated away from it. This means that the leg on one side will swing when the leg on the other side steps, and that the leg steps will be synchronized to the trunk movement in a kinematically efficient way. This behavior is efficient because a stepping leg can push the trunk segment forward more easily when the trunk segment is rotated towards the leg. 

Modulation by the contraction detector produces an in-phase gait pattern, by pulling the legs on both sides of the trunk segment to stance phase when the segment is pulled forward. In equations \ref{eqn:lcT} and \ref{eqn:lcT_2}, we can see that the modulation has the same sign for both legs. This means that this modulation will pull both leg phases to stance phase at the same time, and to swing phase at the same time, making the legs move in sync with each other. The legs are pulled to stance phase when the muscles in front of them are contracted, which in turn makes the body controller contract the muscles below the legs. With the delay introduced by the parameter $c$, this produces a peristaltic wave of contraction that propagates down the trunk.

The body controller contracts muscles in the trunk beneath the legs that are in contact with the ground. When the angle detector has pulled the left and right legs to opposite phases, only the trunk muscle on one side will contract at a time, reinforcing the rotating/undulatory motion of the trunk. When the contraction detector has pulled both legs to the same phase, the trunk muscles will contract on both sides at the same time, reinforcing the peristaltic motion of the trunk. 

\paragraph{Foot controller}
The leg controller moves the outer segment of the legs around the roll axis of the centipede.
The control of the foot is based on the phase of the corresponding leg controller. The foot moves in sync with the leg. The angle of the foot joint, $F$, is determined with the following equation:

\begin{equation}\label{eqn:fc}
F^P_i = \alpha_F \sin(\theta^P_i)
\end{equation}

where $P$ can be either $L$ or $R$, which are the left and right side of the model, $i$ is the segment number indexed from tail to head, $\theta$ is the phase of corresponding leg controller, and $\alpha_{F}$ is the amplitude of the foot.

\paragraph{Controller parameters}
The parameters we used for the controller are summarized in the table below.
The parameters were found through hand tuning. 

\begin{center}
\label{tab:controller_parameters}
\begin{tabular}{|l|r||l|r||l|r|}
    \hline
    Body &  & Leg &  & Foot &  \\
    \hline
    $l_n$ & $1$ & $\alpha_{LEG}$ & $20^{\circ}$ & $\alpha_F$ & $20^{\circ}$ \\
    $\gamma$ & $0.005$ & $\omega$ & $0.02\pi$ & & \\
    & & $\sigma_A$ & $0.09$ & & \\
    & & $\sigma_T$ & $0.076$ & & \\
    & & $c$ & $0.5$ & & \\
    \hline
\end{tabular}
\end{center}

\subsection{Pattern detection}
\label{sec:patterndetection}
Throughout the experiments we use several measurements to analyze and categorise the gait patterns of the centipede model. All the measurements, and the categorisation method, are described below. 

\paragraph{Distance moved}
The distance moved is measured as the euclidean distance between the position of the head segment at the beginning, and end, of the simulation.

\begin{figure}[tp]
    \label{fig:show_patterns}
    \centering
    \includegraphics[width=\textwidth]{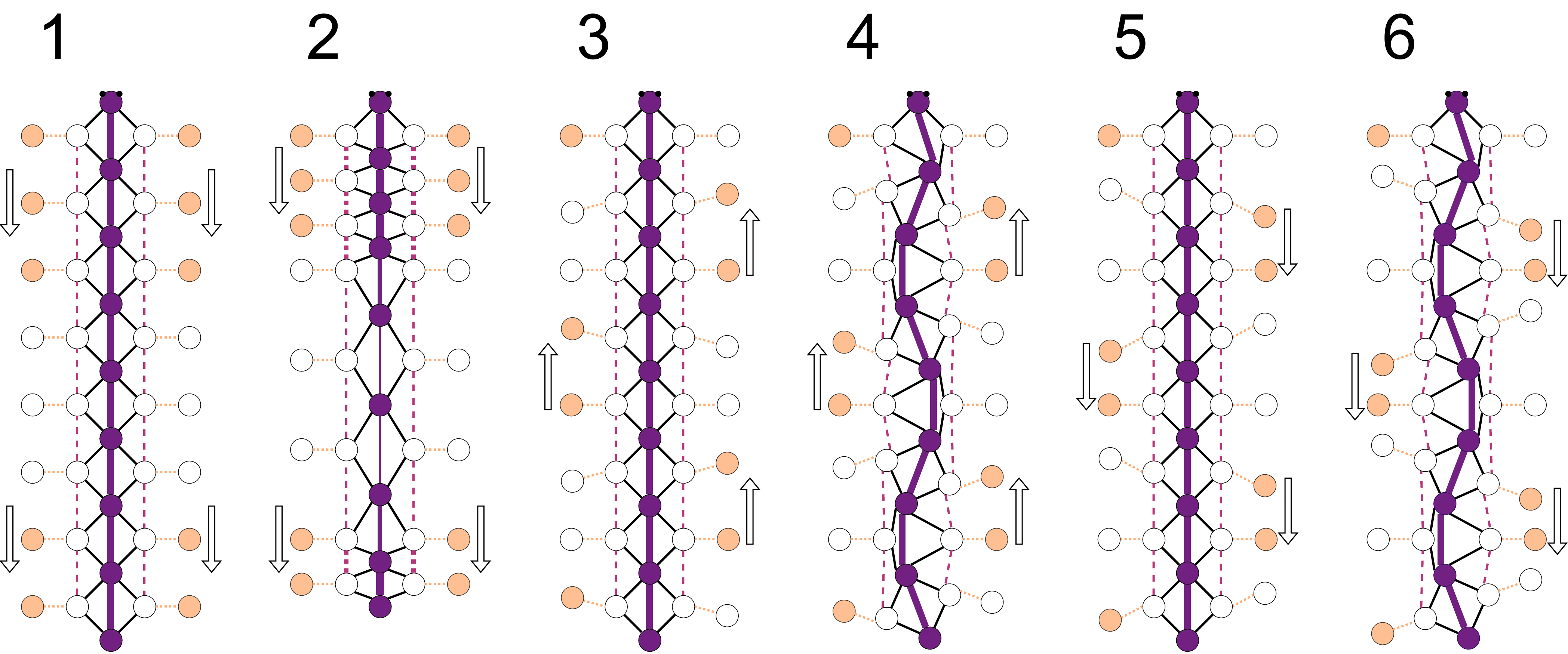} 
    \caption{The six patterns we detect to categorize the gaits, 1) in-phase gait, 2) peristaltic gait, 3) low undulation direct gait, 4) high undulation direct gait, 5) low undulation retrograde gait, and 6) high undulation retrograde gait. The retrograde gaits produce a visual pattern where it looks like clusters of legs are moving downwards towards the tail. The direct wave gaits produce the opposite effect where it looks like the clusters are moving upwards towards the head. This is denoted by the arrows. The in-phase and peristaltic gaits (1 and 2) usually exhibit a retrograde wave.}
    \label{fig:gaits}
\end{figure}

\paragraph{Ipsilateral phase difference}
The ipsilateral phase difference is the difference in phase of a leg and the leg in front of it. The ipsilateral phase difference determines the type of wave, direct or retrograde (see figure \ref{fig:show_patterns} for a description of the gaits). For segment number i the equation is: 

\begin{align}\label{eqn:tpd}
    & IPD_i^L = \arctan(\sin(\theta_i^L - \theta_{i+1}^L),\cos(\theta_i^L - \theta_{i+1}^L))\\
    & IPD_i^R = \arctan(\sin(\theta_i^R - \theta_{i+1}^R),\cos(\theta_i^R - \theta_{i+1}^R))\\
    & IPD_i = \frac{IPD_i^{L} + IPD_i^{R}}{2}
\end{align}

where $\theta_i^L$ is the phase of the left leg of the $i$th segment and $\theta_i^R$ is the phase of the right leg of the $i$th segment.

\paragraph{Contralateral phase difference}
The contralateral phase difference is the difference in phase between the left and right leg of a segment. The contralateral phase difference determines whether the legs on either side of the body are in phase. Typically the legs are not in  phase during undulatory locomotion, and are in phase during peristaltic locomotion. For segment number i the equation is:

\begin{equation}\label{eqn:lpd}
CPD_i = \arctan(\sin(\theta_i^L - \theta_i^R),\cos(\theta_i^L - \theta_i^R))
\end{equation}

where $\theta_i^L$ is the phase of the left leg of the $i$th segment and $\theta_i^R$ is the phase of the right leg of the $i$th segment.

\paragraph{Undulation}
The undulation is a measure of the curvature of the trunk. The curvature of the trunk is measured by finding the angle between a trunk segment and the trunk segment in front of it. An increased curvature of the trunk, which is synchronized with the leg movement, leads to an increased stride length for the leg. This often leads to faster locomotion. For segment number i the equation is:

\begin{equation}\label{eqn:ua}
a = F_{i-1}^{pos} - B_{i-1}^{pos}
\end{equation}
\begin{equation}\label{eqn:ub}
b = F_i^{pos} - B_i^{pos}
\end{equation}
\begin{equation}\label{eqn:u}
U_i = |\arctan(||a\times b||,a\cdot b)|
\end{equation}

where $F_i^{pos}$ is the position of the front mass point in segment i, and  $B_i^{pos}$ is the position of the back mass point in segment i.

\paragraph{Segment length}
The segment length is measured to determine how compressed the trunk is during a peristaltic gait. For segment number i the equation is:

\begin{equation}\label{eqn:l}
BL_i = ||F_i^{pos} - B_i^{pos}||
\end{equation}

where $F_i^{pos}$ is the position of the front mass point in segment i, and  $B_i^{pos}$ is the position of the back mass point in segment i.

\paragraph{Categorisation of gait patterns}
We categorise the gaits into 6 patterns. Two of the patterns exhibit peristalsis, where the trunk segments contract on both sides simultaneously. Four of the patterns exhibit undulation, where the trunk segment contract alternatingly on the left and right side. The patterns are 1) a contralaterally in-phase pattern with low peristalsis, we call it the in-phase gait, 2) a contralaterally in-phase pattern with high peristalsis, we call it the peristaltic gait, 3) a pattern with an undulatory wave passing from tail to head, where the amplitude of the undulation is very small, we call it a low undulation direct gait, 4) a pattern with an undulatory wave passing from tail to head, where the amplitude of the undulation is high, we call it a high undulation direct gait, 5) a pattern with an undulatory wave passing the opposite way from the direct gaits, from head to tail, where the amplitude of the undulation is very small, we call it a low undulation retrograde gait, 6) a pattern with an undulatory wave passing from head to tail, where the amplitude of the undulation is high, we call it a high undulation retrograde gait. All the gaits are depicted in Figure \ref{fig:gaits}. We call gaits 1 and 2 peristalsis-type gaits, and 3 to 6 undulation-type gaits.

\begin{figure}[tp]
    \centering
    \includegraphics[width=\textwidth]{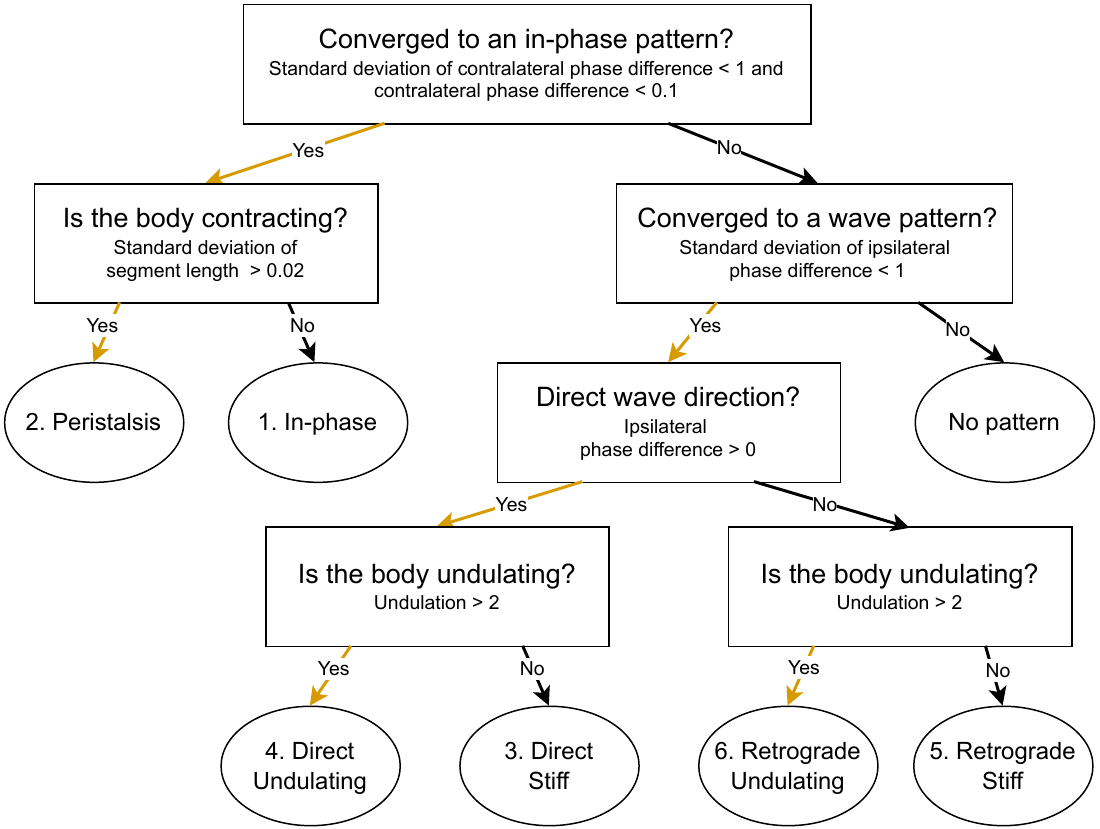} 
    \caption{Flowchart showing the criteria for detecting a pattern in a centipede gait.}
    \label{fig:flowchart}
\end{figure}

The analysis measurements described at the beginning of this section are defined for a single segment. When determining which pattern appeared, we find the average and standard deviation of these measurements over all segments. To reach the single values used in the classification, we take the average of each measure over the last 200 frames of the simulation. 

To classify a pattern, a series of checks are performed (see Figure \ref{fig:flowchart} for a flowchart summarizing the process). We first determine whether the pattern is contralaterally in phase. 
A gait is considered as contralaterally in phase if the contralateral phase difference is less than 0.6, and the standard deviation of the contralateral phase difference is less than 0.7. An in-phase pattern is considered to have high peristalsis if the standard deviation of the segment length is greater than 0.02.
If the pattern was not classified as an in-phase gait, we next determine whether it can be categorised as a wave gait. A pattern is considered a wave gait if the standard deviation of the ipsilateral phase difference is smaller than 0.7.
The direction of the undulatory wave, direct or retrograde, is also determined by the ipsilateral phase difference. If the ipsilateral phase difference is greater than zero the wave is direct, if it is negative the wave is retrograde. Both the direct and retrograde wave patterns are considered to have high undulation if the undulation is above 2. If the pattern was not categorised by any of the criteria above it is considered a gait that has not converged.

Although we do not classify the contralaterally in-phase patterns (1 and 2) into direct or retrograde waves, we observed that these patterns almost always converge to a retrograde wave, where the ipsilateral phase difference is less than zero. 

\section{Results}
\begin{figure}[tp]
    \centering
    \includegraphics[width=\textwidth]{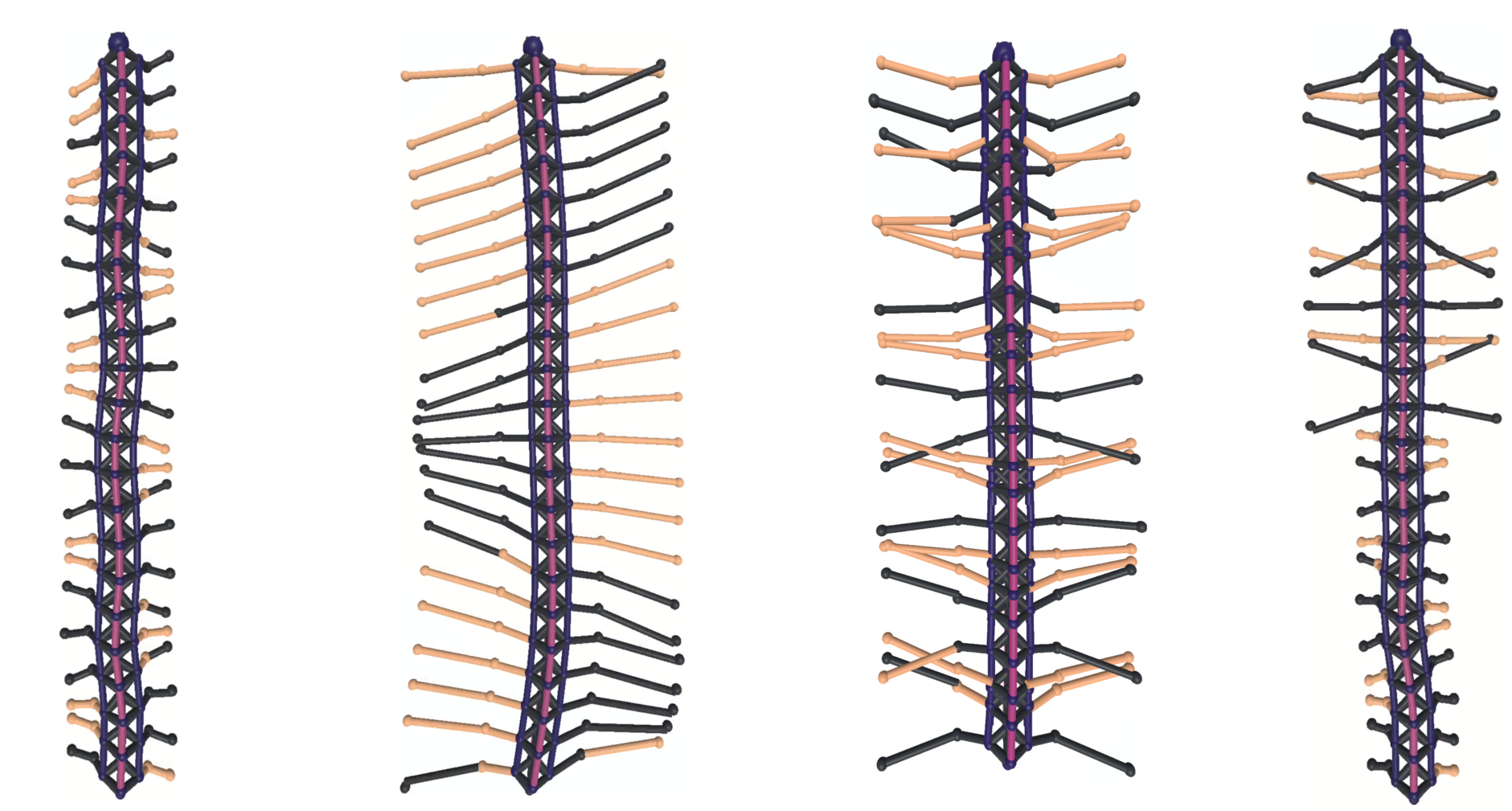}
    \caption{Examples of gaits observed in the centipede model: a) an undulatory retrograde wave gait, b) a direct wave gait, c) an in-phase gait with peristalsis, and d) a mix of an in-phase gait and an undulatory retrograde wave gait.}
    \label{fig:four_patterns}
\end{figure}

We conducted four experiments. In the three first experiments the three variants of the controller, A, C and A+C, are compared quantitatively. The first and second experiment is in the walking scenario, and looks at adaptation to changes in leg length and mass respectively. The third experiment is in the climbing scenario, and looks at adaptation to the incline of the pole. The last experiment tests the A+C controller on a morphology with mixed leg lengths, in the walking scenario, and is a qualitative analysis of the emerging gait. Images of four of the gaits that appeared throughout the experiments can be seen in figure \ref{fig:four_patterns}.

All data plotted is the result of ten simulation trials. When categorising the patterns, the pattern that that appeared most frequently during the ten simulation trials is reported.
In the time series the average of the ten simulation trials is reported. Note that in the time series plots the standard deviation is not the standard deviation of the ten trials, but the standard deviation of the centipede segments averaged over the ten simulation trials.

\subsection{Walking scenario: Leg length}

\begin{figure}[tp]
    \centering
    \includegraphics[width=\textwidth]{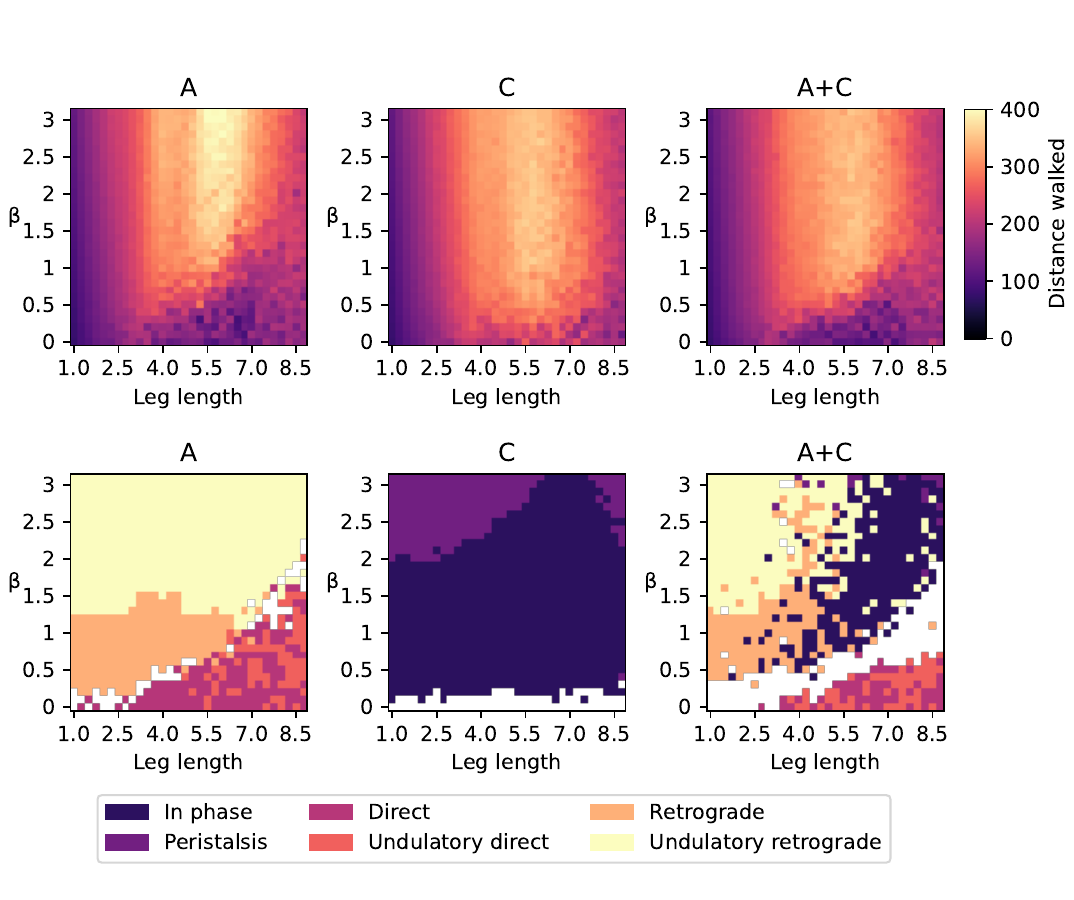} 
    \caption{The results of simulating the centipede model in the walking environment with different leg lengths. The beta parameter of the controller and the leg length was varied as denoted by the axes. The top row shows the distance the centipede model walked. The bottom row shows which gait pattern appeared most frequently in each condition. A white cell means the model did not converge to any pattern. Each column is for one of the three controller variants.}
    \label{fig:exp1}
\end{figure}

In this experiment (see Figure~\ref{fig:exp1}) we compare the three controller variants, A, C and A+C in the walking scenario, while varying the leg length and the $\beta$ parameter. We observe that the centipede model walks faster with the A variant in most situations. The only exception is for the combination of long legs and low $\beta$. In this situation the C variant walks faster. The optimal leg length for high walking speed seems to lie around 5.5 to 6.5.

For the A controller variant the four undulation-type patterns appear. For the C variant the two peristalsis-type patterns appear. This is expected as the A component of the controller is designed to apply opposite modulation to the left and right leg of a segment, while the C component applies the same modulation to both sides. For the A controller variant we see that the direct pattern appears for low $\beta$ values, and up to a $\beta$ value of $1.5$ when the legs are long. The undulation of the direct wave gait seems to be larger when legs are longer. The A+C variant mainly exhibits the same patterns as the A variant, except for the combination of long legs and high beta, where the in-phase pattern appears.
While the A and C variants have clear areas where each pattern appears, the A+C variant seems to have less defined boundaries.
The combination of the noisy boundaries and lower walk speed for the A+C variant suggests that the C component is not beneficial for the centipede model while it is walking.
For both the A and A+C variants we observe a drastically lower walking speed when the direct wave patterns appear. 

With this experiment we also display that our controller is capable of mode switching. Looking at Figure~\ref{fig:exp1}, controller variant A, we see that in the row with a $\beta$ value of 0.5, a retrograde gait appears for short legs, a direct wave gait appears for medium-long legs, and a undulatory direct wave gait appears for long legs. The centipede model switches from one type of gait to another, solely based on changes in the morphology. For controller variant A+C mode switching happens for high $\beta$ values, where an undulatory retrograde pattern appears for short legs, while a contralaterally in phase pattern appears for long legs.

Looking at the bottom rows of the graphs in figure \ref{fig:exp1}, where $\beta$ is $0$, we can see that even when the body contraction amplitude zero, meaning that the body controller is turned off, and the trunk does not respond to the touch information from the feet, the model is still able to converge to a gait pattern. This is likely because the momentum from the feet causes the trunk to bend. The legs then synchronize to the trunk bend, which is caused only by the implicit dynamics of the model. When allowed to synchronize in this way, the model converges to a direct wave gait, for the A and A+C variants of the controller. 

When the legs of the centipede model are longer, their momentum becomes larger, which might explain why the undulation of the direct wave becomes larger as the leg length increases. This can also explain why there is a mode switch from retrograde to a direct wave when leg length increases. The retrograde wave gaits appear when the contractions in the trunk, caused by the body controller, become large enough to overpower the implicit direct wave. In the white band, located in between the section with retrograde patterns and the section with direct patters, the retrograde and direct wave cancel each other out. In this band the model is not capable of converging to a gait. In the A+C variant of the controller the in-phase pattern appears when the legs are very long. This might be because it is more difficult to keep the long legs above the ground. If both feet touch the ground at the same time, and $\beta$ is high, the contraction detector will detect contraction of the trunk, reinforcing the peristalsis-type gaits.

In the A+C variant of the controller the A (angle detector) and C (contraction detector) components become two basins of attraction. When the proprioceptive feedback of $A^D$ is high $C^D$ will be low, and when the feedback of $C^D$ is high $A^D$ will be low. This causes the model to converge either a peristaltic or undulatory mode, based on bias from initial conditions, environment or morphology. The extension from A to A+C did not increase the walking speed for the morphologies that we tested, but it has the benefit that it adds additional modes of locomotion that can be better in other scenarios. We demonstrate that in the climbing scenario in experiment 3, where the C component benefits the model.

\subsection{Walking scenario: Mass}

\begin{figure}[tp]
    \centering
    \includegraphics[width=\textwidth]{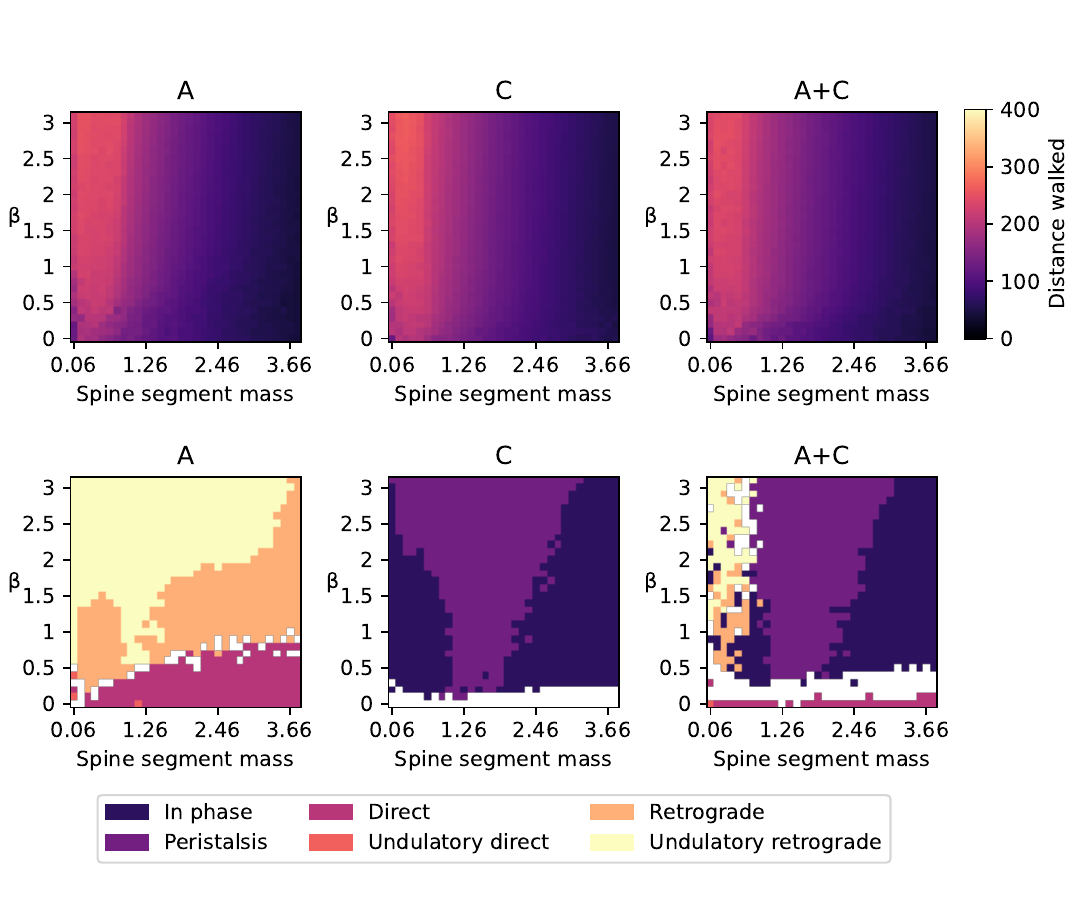} 
    \caption{The results of simulating the centipede model in the walking environment with different trunk masses. The beta parameter of the controller and the trunk mass was varied as denoted by the axes. The top row shows the distance the centipede model walked. The bottom row shows which gait pattern appeared most frequently in each condition. A white cell means the model did not converge to any pattern. Each column is for one of the three controller variants.}
    \label{fig:exp4}
\end{figure}

In this experiment (see Figure~\ref{fig:exp4}) we compare the three controller variants, A, C and A+C in the walking scenario, while varying the mass of the trunk and the $\beta$ parameter. Similarly to when varying the leg length, the performance of the three variants is similar, but the A variant has a slightly higher walk speed for low mass.

All four undulation-type patterns appeared for controller variant A, and the two peristalsis-type patterns appeared for variant T. In variant A, a mass increase reduces the amount of undulation and, for $\beta$ values around $0.5$, induces a mode switch from a retrograde to a direct wave gait. In the C variant there is a band of mass values where the centipede model exhibits high peristalsis, for high $\beta$ values this band becomes wider. For the A+C controller variant all six patterns appeared. The peristalsis-type patterns appear in most cases, but for low mass the undulatory retrograde wave patterns appear, and for low $\beta$ values the undulatory direct wave pattern appear. For most $\beta$ values we observe a mode switch from a retrograde pattern for low mass to a peristaltic pattern for medium mass, and then to an in-phase pattern for high mass. 

The higher trunk mass makes it more difficult for the legs to move the trunk segments. This may be the reason for the mode switch from retrograde gaits to peristalsis type gaits as the mass increases. The retrograde pattern only uses one leg at a time to push the trunk segment, while the peristalsis-type patterns use both legs simultaneously, making it easier to move the heavy trunk segments with a peristalsis-type pattern. This could make the model more likely to converge to a peristalsis type gait.

\subsection{Climbing scenario}
\begin{figure}[tp]
    \centering
    \includegraphics[width=\textwidth]{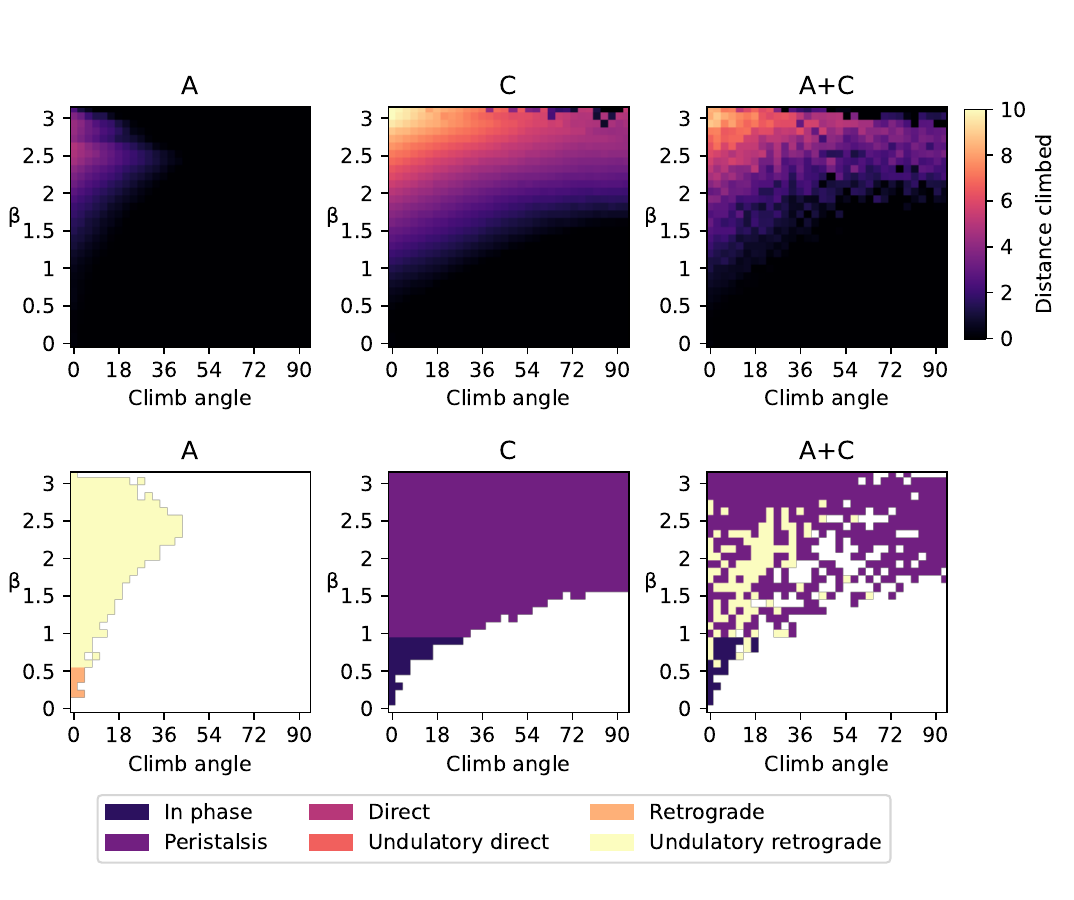} 
    \caption{The results of simulating the centipede model in the climbing environment.  The beta parameter of the controller and the pole angle was varied as denoted by the axes. The top row shows the distance the centipede model climbed. The bottom row shows which gait pattern appeared most frequently in each condition. A white cell means the model did not converge to any pattern, or fell of the pole. Each column is for one of the three controller variants.}
    \label{fig:exp2}
\end{figure}

In the climbing scenario the C controller variant had the highest performance (see Figure \ref{fig:exp2}), showing that the C component is beneficial for climbing. The A+C variant has slightly lower performance than the C variant, likely due to noise from the A component, but it still performs significantly better than the A variant. When the $\beta$ value is low the centipede model is not capable of climbing with any of the three controller variants. This is likely because the leg movement contributes very little to the upwards momentum when they are folded around the pole. The model therefore has to rely on trunk contraction to create the upwards momentum. Since the trunk contracts very little for low $\beta$ values it becomes impossible to climb with low $\beta$.

As in the walking scenario the A controller variant only produced undulation-type gaits, while the C controller variant only produced peristalsis-type gaits. None of the variants produced direct wave gaits. The A+C variant exhibits both retrograde wave gaits and the peristalsis gait. The peristalsis gait becomes more prominent for steeper pole angles, and high $\beta$. This could be because the trunk straightens out as the centipede grips the pole to hold on and keep from falling.  When a trunk segment is rotated , as in an undulatory gait, the segment is only supported on one side, while when the legs are contralaterally in phase the body weight is supported on both sides.
 
By looking at both Figure \ref{fig:exp1} and Figure \ref{fig:exp2}, we demonstrate automatic gait switching based on environment changes for the A+C variant. The centipede model used in the climbing scenario always has a leg length of 1.5. In the column for leg length 1.5 (3rd column from the left), in Figure \ref{fig:exp1}, we see only retrograde wave gaits as the centipede model is walking. The same model, with the same controller parameters, displays the peristalsis gait for many different pole inclines in the climbing scenario.

\subsection{Multi-pattern morphologies}

\begin{figure}[tp]
    \centering
    \includegraphics[width=\textwidth]{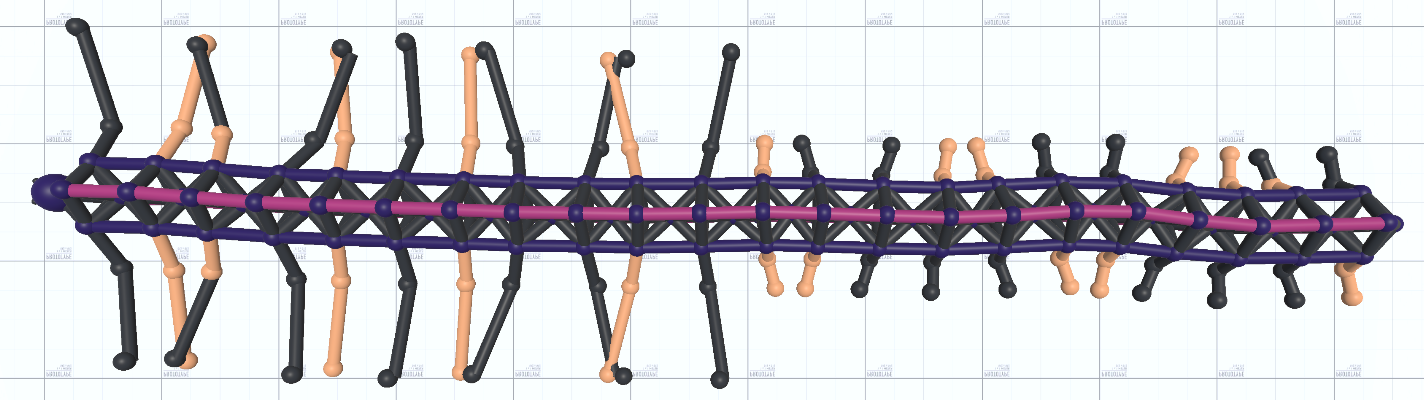} 
    \includegraphics[width=\textwidth]{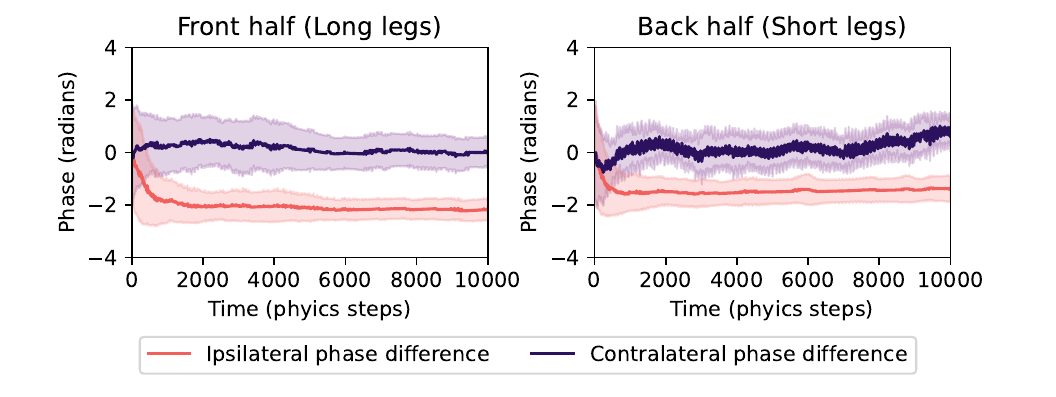} 
    \caption{The ipsilateral and contralateral phase differences, over time, for the front and back part of a mixed morphology centipede. The front section has converged to the in-phase pattern, while the back part has converged to an undulatory retrograde pattern.}
    \label{fig:exp5}
\end{figure}

In the final experiment we qualitatively analyse one centipede morphology. Looking at figure \ref{fig:exp5} we see a centipede model with a mixed leg length morphology. The front half of the centipede has long legs, while the back half has short legs, and $\beta$ is set to 3. When the model had this mixed morphology, the two sections converge to different patterns. The front half has converged to an in-phase pattern (the contralateral phase difference is close to 0) while the back half has converged to an undulatory retrograde pattern (the contralateral phase difference is noisy, while the ipsilateral phase difference is close to -1.5). The convergence to different patterns in different parts of the model is possible because the controller is fully decentralized and reacts to local forces in the model. 

When cutting this centipede in two, both sections walk with a retrograde wave gait. We can see this by looking at the 1st and 9th column (leg length 1 and 3) of the bottom right graph in figure \ref{fig:exp1}. These two columns represent a centipede with respectively only the long or short legs of this model, and about half the number of segments. Both of these converged to retrograde wave gaits. In the mixed morphology the forces experienced by the front half of the model is different because of the added back half with a different morphology. The difference in forces could be caused by the section with short legs walking slower than the section with long legs, and pulling the front half backwards, resulting in the mixed gait.

\section{Discussion}

In this paper we presented a controller for centipede locomotion. The behaviour of the controller was explored through simulation of a centipede-like virtual creature. We showed that the controller has a large behaviour diversity despite being a simple mathematical model. The controller exhibits six different locomotion modes, two gaits with peristalsis in the trunk, two gaits with direct undulatory waves, and two gaits with retrograde undulatory waves.
The model automatically switches between the gaits depending on features of the morphology and environment. The changes in morphology and environment changes the dynamics of the system, which causes the controller to converge to different gait patterns. In our experiments we show mode switches dependent on 1) the body contraction amplitude, 2) the leg length, 3) the trunk weight, and 4) incline in the environment. The controller also shows flexibility, and can adapt to local morphological features, converging to different patterns in different parts of the centipede body.

A novel feature of our controller was a mechanism for generating peristaltic movement in the trunk. When the centipede model was moved from a walking scenario, to a scenario where it was climbing up a pole, the peristaltic movement automatically emerged.
The centipede model had a significantly higher climbing speed, and ability, when using the peristaltic gait compared to the undulatory gait. The peristalsis in the trunk seemed to only be beneficial when climbing, as it did not grant any benefits when the model was walking.
The undulatory gaits were most prominent when the centipede model was walking and had low trunk mass, and the centipede models with the highest walking speed were among those that exhibited the retrograde wave gait. 
When the model converged to a direct wave gait there was a drop in walk speed. This may be because the natural undulatory wave that appears due to the momentum of the legs is kinematically inefficient when combined with the direct wave leg movement. This is also supported in biology, as it is believed that some insects, that have direct wave gait pattern, have very stiff bodies to counteract this momentum-induced undulation \citep{manton1977arthropoda}. Although the direct wave gaits seem less efficient that the retrograde wave gaits in the context of our model, we believe they may have other benefits not directly related to walking speed,
such as energy conservation. 
The full controller, which exhibits all six patterns, is more versatile in adaptation to the environment. However, this comes with a tradeoff in maximum walking speed. Noise from the mechanism producing the peristaltic patterns reduces the walking speed in some of the scenarios. This is most visible in the scenarios where the retrograde wave gait is the most efficient option. In future work we would like to investigate how to either separate these two components, or enhance the convergence to a pattern to reduce this noise. We would also like to compare this controller to other more traditional controllers typically used in robotics.

A limitation of our work is that the controller is only tested on centipede-like morphologies, and only in simulation. In future work we would like to test the controller on more diverse morphologies, and expand it to automatically adapt to these morphologies with new and relevant behaviors. This could include balancing mechanisms for walking on few legs, more advanced control of the trunk contraction for legless gaits, or phase reversal to produce kinematically efficient direct wave gaits. 
A controller with a large behaviour diversity, and adaptability to morphology and environment, would be highly useful in the design of physical robots, as it could offer versatility, robustness, and quick adaptation to damage.
Extending the controller to accommodate even more diverse morphologies would greatly extend the potential application domains, and should be studied in future work.
The current centipede model may need some modifications before it can be reproduced in the real world. 
As we have focused on implementing a virtual model with high configuration and actuation flexibility for testing the controller approach, the fidelity of the simulation model may not in all cases correspond to real world constraints. Therefore, we expect further work on simulator tuning and adjusting to a real-world robotic configuration to be necessary. 
Although there are solutions for linear actuators that could be adopted, the linear actuators in the trunk are less straightforward to implement than the rotational actuators, and various designs may need to be considered.

An evolutionary time scale could also be added to explore how controllers and morphologies develop together to adapt to environments. As the controller adapts to the morphology during the evaluation phase, it does not need any re-training to adapt to new morphologies. This could speed up the evolutionary process as new morphologies could be evaluated quickly. The controller may also have potential to enhance adaptation of the morphology to the environment, as the controller will not overfit to a single morphology. Optimisation of morphology through evolutionary algorithms could also be a good testbench to compare this controller to more traditional robot controllers with less domain knowledge, which will be an important consideration in future work.

\section{Conclusion}
We have presented a controller for centipede-like locomotion. We showed through simulations on a centipede model that the controller can exhibit multiple modes of locomotion. The modes include variations of undulatory and peristaltic trunk motion, combined with legged motion that synchronizes to the movement of the trunk. The centipede model requires no training or parameter adaptation, and automatically converges to one of the locomotion modes in the first part of the evaluation in the simulation environment. The centipede locomotion automatically adapted to different morphologies and environments by switching between the different locomotion modes. The peristaltic locomotion modes appeared during climbing, where they proved more efficient at creating upward speed than the undulatory patterns. 

This paper encompassed centipede-like morphologies, and two environments, but we plan to extend the controller to produce sensible behaviors in even more varied morphologies and environments. Given the versatiliy in adaptation to morphology and environments, we envision that the presented adaptive controller approach could provide insights in studies on simultaneous evolution of morphology and control, and be adaptive to damage in the morphology on a hand designed robot.

\section{Data availability}
The code for the project, and the executable for the simulator, can be found at \url{https://github.com/EmmaStensby/myriapod}. This can be used to reproduce the data.

\section{Acknowledgements}
This work is partially supported by The Research Council of Norway (RCN) as a part of the project Collaboration on Intelligent Machines (COINMAC-2) project, under grant agreement no. 309869, and through its Centres of Excellence scheme, RITMO, with project no. 262762.

\printbibliography

\end{document}